%% file: main.tex
\documentclass[sigconf,nonacm]{acmart}
\AtBeginDocument{%
  }

\setcopyright{acmlicensed}
\copyrightyear{2018}
\acmYear{2018}
\acmDOI{XXXXXXX.XXXXXXX}
\acmConference[Conference acronym 'XX]{Make sure to enter the correct
  conference title from your rights confirmation email}{June 03--05,
  2018}{Woodstock, NY}
\acmISBN{978-1-4503-XXXX-X/2018/06}

\usepackage{url}
\usepackage{tikz}
\usetikzlibrary{arrows.meta, positioning, shapes.geometric, decorations.pathreplacing}
\usepackage{amsmath}
\usepackage{amsfonts}
\usepackage{subcaption}
\usepackage{graphicx} % Required for inserting images
\usepackage{multicol}
\usepackage{multirow}
\usepackage{booktabs}
\usepackage{adjustbox}
\usepackage{hyperref}
\usepackage{tabularx}  
\usepackage{array} 
\usepackage{macros}

\begin{document}

% \title{\defensename: A Principled Defense Against Backdoors in Federated Learning}
\title{\defensename: Toward a Theory of Backdoors in Federated Learning}

\author{Lucas Fenaux}
\email{lucas.fenaux@uwaterloo.ca}
\orcid{0000-0001-8008-4037}
\affiliation{%
  \institution{University of Waterloo}
  \city{Waterloo}
  \state{Ontario}
  \country{Canada}
}

\author{Zheng Wang}
\orcid{0009-0006-0152-4078}
\affiliation{%
  \institution{University of Waterloo}
  \city{Waterloo}
  \state{Ontario}
  \country{Canada}
}

\author{Jacob Yan}
\orcid{0009-0005-5396-8626}
\affiliation{%
  \institution{University of Waterloo}
  \city{Waterloo}
  \state{Ontario}
  \country{Canada}
}

\author{Nathan Chung}
\orcid{0009-0008-6150-9501}
\affiliation{%
  \institution{University of Waterloo}
  \city{Waterloo}
  \state{Ontario}
  \country{Canada}
}

\author{Florian Kerschbaum}
\orcid{0000-0003-4288-2286}
\affiliation{%
  \institution{University of Waterloo}
  \city{Waterloo}
  \state{Ontario}
  \country{Canada}
}
\renewcommand{\shortauthors}{Fenaux et al.}


\newcommand{\todo}[1]{{\color{green}[TODO]~{#1}~[/TODO]}}

\newcolumntype{L}{>{\raggedright\arraybackslash}X}

% \textbf{\del{
% WRITING NOTES FOR CONSISTENCY: WORDS WE USE:
% \begin{itemize}
%     \item aggregator  (not server)
%     \item benign (not clean)
%     \item attacker-controlled behavior (not malicious behavior)
%     \item malicious is only for clients and updates, nothing else
% \end{itemize}}}
% 

\begin{abstract}

Federated Learning (FL) enables distributed model training but is vulnerable to backdoor attacks, where malicious clients embed attacker-controlled behaviors into the global model. Existing defenses fail against adaptive adversaries. In this paper, we present \textbf{\defensename}, a principled theoretical framework that categorizes backdoors by the deviation, $\delta$, of their updates to the mean of the updates. We identify two fundamental defense types: \textbf{Type 1 (The Anvil)}, comprising outlier detection and robust aggregation effective against large-deviation attacks, and \textbf{Type 2 (The Hammer)}, consisting of removal-based defenses effective against small-deviation attacks. We demonstrate that defenses of a single type and non-principled combined defenses inherently leave an exploitable gap for adaptive attackers. To bridge this gap, we propose the principled combination of Type 1 and Type 2 defenses. We evaluate our framework against a new, worst-case, full-information adaptive adversary that knows the benign updates, the aggregation algorithm, and its parameters, and yet this adversary fails against our combined defenses. Our empirical evaluation across various datasets and settings shows that single-typed and non-principled combined defenses are easily broken, often by a single malicious client. In contrast, our best combined defense variants, $HA_{Flame}^{CSFT}$, $HA_{Krum}^{CSFT}$, and $HA_{Multi-Metrics}^{CSFT}$, remain undefeated even in the most adversarial settings. Our results provide a principled approach for research on backdoors in federated learning.

\end{abstract}

 %  <concept_id>00000000.00000000.00000000</concept_id>
 %  <concept_desc>Do Not Use This Code, Generate the Correct Terms for Your Paper</concept_desc>
 %  <concept_significance>300</concept_significance>
 %  <concept_id>00000000.00000000.00000000</concept_id>
 %  <concept_desc>Do Not Use This Code, Generate the Correct Terms for Your Paper</concept_desc>
 %  <concept_significance>100</concept_significance>
 %  <concept_id>00000000.00000000.00000000</concept_id>
 %  <concept_desc>Do Not Use This Code, Generate the Correct Terms for Your Paper</concept_desc>
 %  <concept_significance>100</concept_significance>


\keywords{Federated Learning, Machine Learning Security, Backdoors}

\maketitle

\section{Introduction}
% Intro FL and backdoors
Federated Learning \cite{mcmahan2017communication} is a technique for efficient distributed learning in which data resides with the data sources, and only model updates are aggregated to a central model.
Although this distribution of data sources enhances privacy, it does not guarantee it, and it creates opportunities for malicious clients to attack.
One example of such an attack is backdooring \cite{gu2019badnets,zhuang2023backdoor,liu2024beyond,nguyen2023iba,xie2019dba,zhang2022neurotoxin,bagdasaryan2020backdoor,gong2022coordinated,zhang2020poisongan,wang2020attack,bhagoji2019analyzing,zhou2021deep,baruch2019little,li20233dfed,zhang2023a3fl,dai2023chameleon,xu2022more,liu2023facilitating,zhao2022defeat,salem2022dynamic,sun2022semi,chang2024fedtrojan,wang2025sba}, where clients embed attacker-controlled behaviors into the model, such as a trigger that forces any image to be classified as a chosen class.
Backdoor attacks have been shown to pose a significant threat in centralized machine learning \cite{gu2019badnets}. The emergence of federated learning has introduced new attack vectors for backdoor attacks. Through malicious clients, adversaries can embed malicious behaviors into the global model. 

% Intro FL defenses
Many diverse defenses have been developed over the years \cite{nguyen2022flame,xu2025detecting,minsker2023efficient,blanchard2017machine,kumari2023baybfed,fereidooni2023freqfed,lyu2023poisoning,fang2023vulnerability,li2024darkfed,huang2025scope,yu2023g,munoz2019byzantine,shen2016auror,fung2018mitigating,sattler2020byzantine,tolpegin2020data,bagdasaryan2019differential,preuveneers2018chained,nguyen2019diot,zhang2022fldetector,yin2018byzantine,guerraoui2018hidden,ozdayi2021defending,bernstein2018signsgd,rieger2022deepsight,xie2021crfl,sun2021fl,wu2020mitigating,liu2018fine,sun2019can,li2020learning,wu2022federated,cao2019understanding,chen2017distributed,pillutla2022robust} to prevent these attacks in the federated setting. These defenses can be separated into three categories depending on when they appear in the federated pipeline: pre-aggregation, in-aggregation, and post-aggregation \cite{nguyen2024backdoor}. They can also be grouped by their underlying approach: outlier-detection-based, robust-aggregation-based, or removal-based. Outlier detection defenses leverage differences between adversarial and benign updates to distinguish between them. Robust-aggregation defenses limit the impact of any update, thereby limiting the adversary's capabilities to insert a backdoor. Finally, removal defenses make minor modifications to the model in order to remove the embedded backdoor.

% Core observation
Crucially, we observe that different defense categories are effective against distinct types of backdoor attacks. Intuitively, backdoored model updates sent by malicious clients that deviate further from benign updates will be detected more easily by outlier detection defenses. Likewise, since robust-aggregation defenses bound the impact of any single update, beyond a certain threshold, further deviations have no impact, meaning that a small deviation attack has the same impact as a large deviation attack. However, removal defenses have more difficulty removing these updates with large deviations without damaging the model's benign performance, but they can easily remove attacks with small deviations. 

% Intro to our theory
From our observations, we reason about backdoors in terms of their expected update deviation $\delta$ to build a principled theoretical framework for federated learning backdoors: \textbf{\defensename}. We categorize defenses into two types: \textbf{Type 1 defenses} (the Anvil: outlier detection and robust aggregation), which are effective against large deviation attacks, and \textbf{Type 2 defenses} (the Hammer: removal), which are effective against small deviation attacks. From our categorization, it follows that defenses of a single type cannot be robust without drastically affecting model utility, as they fundamentally leave a gap for an adaptive attacker to exploit. Hence, we argue for combining defenses using our principled theoretical framework. By combining non-interfering Type 1 and Type 2 defenses, we can reduce the exploitable gap for the attacker, or even eliminate it. 

% Explain our threat model
We ground our principled theoretical framework in worst-case assumptions for security to ensure secure deployment of FL in any setting. Hence, we design a new adversary threat model for federated learning: the \textbf{full-information adaptive adversary}. Our full-information adaptive adversary knows: the aggregation algorithm, its parameters, and the benign updates submitted by all other clients before it has to submit its own, for every epoch. Our adversary can freely collude with the other malicious clients it controls and can adapt to the benign updates submitted by the other clients. Mirroring cryptographic scheme design, a defense robust against our full-information adaptive adversary would implicitly be robust against any attacker with more realistic assumptions.

We empirically validate our principled theoretical framework in two steps: (1) We design principled adaptive attacks against 13 single-typed and non-principled combined defenses \cite{blanchard2017machine, sun2019can, minsker2023efficient, huang2023multi, wan2025mars, xu2025identify, mia2025bart, xu2025detecting, fung2018mitigating, liu2018fine, nguyen2022flame} following our theoretical framework. Our adaptive attacks break these defenses with \textbf{at most four} malicious clients out of 20, and \textbf{as few as a single one}; (2) We evaluate our principled combined defense \textbf{\defenseshortname~}against our full-information adaptive adversary and existing popular and state-of-the-art attacks across various data settings. We find that our three best principled combined defense variants, namely \textbf{$\text{HA}_{\text{Flame}}^{\text{CSFT}}$}, \textbf{$\text{HA}_{\text{Krum}}^{\text{CSFT}}$}, and \textbf{$\text{HA}_{\text{Multi-Metrics}}^{\text{CSFT}}$}, \textbf{are undefeated in every setting we evaluate}. Our empirical results support our theory and enable principled research on backdoor attacks in federated learning.

% Summarize our contributions
% We also propose a theory of backdoor attacks and defenses, supported by empirical evidence. From our theory, we derive a principled approach to combining defenses that yields defenses robust in the most challenging threat model to date.
We summarize our contributions as follows:
\begin{itemize}
    \item We present a new principled theory of backdoors in federated learning: \textbf{\defensename}~in Section \ref{sec:principle}. Our theory categorizes defenses into two types: \textbf{Type 1} (Anvil) and \textbf{Type 2} (Hammer), and reveals a fundamental exploitable gap in single-typed defenses and for non-principled combined defenses. We then derive rules for the principled combination of backdoor defenses.
    \item We introduce a new worst-case threat model: \textbf{the full-information adaptive adversary} in Section \ref{sec:motivation} and use it to break single-typed and non-principled combined defenses in Section \ref{sec:adaptive_attacks}.
    \item We empirically evaluate our principled combined defense \textbf{\defenseshortname~}in Section \ref{sec:def_results}. We show that our three best combined defense variants are undefeated in the worst-case threat model, showcasing state-of-the-art performance. Our empirical results provide support for our theory.
\end{itemize}

\section{Background}\label{sec:background}

\noindent \textbf{Notation:}
Let $ w \in \mathbb{R}^k $ represent the set of weights for a model. We use the terms "model" and "weight of a model" interchangeably, as all client models share the same model architecture. Let $ b $ and $ d $ denote datasets, while $ B $ and $ D $ represent distributions. The function $ T $ is a training function that produces a set of weight updates $ u $ based on a given set of weights and a dataset. Let $n$ be the number of clients and $m$ the number of malicious clients.

\subsection{Federated Learning}
Federated Learning (FL) \cite{mcmahan2017communication} is a distributed machine learning approach where multiple clients collaboratively train a global model without centralizing their local datasets. Formally, $n$ clients collaborate to train a shared model by aggregating updates computed locally on distributed datasets ( $d_i \sim D, i \in \{1, \dots, n\}$). These updates are aggregated by the aggregator to create the global model $w$. We detail the training process for any epoch $t$ below; this series of steps is repeated until convergence or a predefined number of epochs ($t_{max}$) is reached.
\begin{enumerate}
    \item \textbf{(Initialization Phase)} The aggregator sends the global model $w_t$ to a subset of clients of size $s$.
    \item \textbf{(Training Phase)} Each client in the subset trains its copy of the global model on its dataset $d_i$ to generate its update $u_i$. The clients then send their update $u_i$ to the aggregator.
    \item \textbf{(Aggregation Phase)} The aggregator aggregates the clients' updates, yielding the new global model $w_{t+1}$.
\end{enumerate}
The standard aggregation rule is averaging (FedAvg \cite{mcmahan2017communication}), given by: $w_{t+1} = w_t + \frac{1}{s}\sum_{i}u_i$.

\subsection{Backdoor Attacks}
Backdoor attacks embed hidden attacker-controlled behaviors within machine learning models during training \cite{gu2019badnets}. Specifically, the attacker aims to alter the model's training such that the resulting model behaves normally on benign inputs but produces attacker-controlled outputs when triggered. To trigger the attacker-controlled behavior, the attacker can insert or exploit a pattern in the training data \cite{gu2019badnets}, which will be learned by the victim model. Depending on the type of trigger, backdoor attacks can be categorized as artificial or semantic \cite{nguyen2024backdoor}.\\
 \noindent\textbf{Artificial Backdoors:} The triggers are artificially added, typically as a distinct pattern or patch (e.g., a white square in the corner of the image).\\
\noindent\textbf{Semantic Backdoors:} The triggers exploit natural features of inputs, associating specific features with attacker-controlled behavior (e.g., green cars).

\paragraph{Attack Objectives.} The attacker's objective is to maximize the Attack Success Rate (ASR), as in the proportion of backdoored inputs that trigger attacker-controlled behavior, while maintaining accuracy on the model's main task, which we will refer to as benign accuracy or accuracy.
Backdoor attacks can be modeled as a multi-objective optimization problem, where the attacker seeks to optimize the model's behavior across both benign and backdoored inputs. For a dataset $d$ of input-label pairs drawn from a benign distribution $D$, a backdoor dataset $b$ drawn from a backdoored distribution $B$, and a loss function $L$ (for example, Cross-Entropy \cite{bridle1990probabilistic} for image classification), the attacker aims to minimize:
\begin{equation}
\min \sum_{(x_i, y_i) \in d} L(x_i, y_i) + \sum_{(x'_i, y'_i) \in b} L(x'_i, y'_i)
\end{equation}

\subsection{Backdoor Defenses}
In scenarios where backdoor attacks pose a threat to the integrity of an FL model, the aggregator may deploy a defense \cite{nguyen2022flame,xu2025detecting,minsker2023efficient,blanchard2017machine,kumari2023baybfed,fereidooni2023freqfed,lyu2023poisoning,fang2023vulnerability,li2024darkfed,huang2025scope,yu2023g,munoz2019byzantine,shen2016auror,fung2018mitigating,sattler2020byzantine,tolpegin2020data,bagdasaryan2019differential,preuveneers2018chained,nguyen2019diot,zhang2022fldetector,yin2018byzantine,guerraoui2018hidden,ozdayi2021defending,bernstein2018signsgd,rieger2022deepsight,xie2021crfl,sun2021fl,wu2020mitigating,liu2018fine,sun2019can,li2020learning,wu2022federated,cao2019understanding,chen2017distributed,pillutla2022robust}. When deploying a defense, the aggregator must balance two objectives: backdoor removal and benign accuracy preservation. Different approaches have been devised over the years and can be categorized as follows:

\noindent \textbf{Robust-aggregation:} limit the impact of any update, thereby limiting the impact of adversarial updates and the adversary's capabilities to insert a backdoor. Targets large deviation magnitudes. %Norm-bounding \cite{sun2019can} and Median-of-Means \cite{minsker2023efficient} are existing robust-aggregation defenses. 
\\
\noindent \textbf{Outlier Detection:} leverage the differences between adversarial and benign updates to distinguish them. Also targets large deviation magnitudes. %Krum \cite{blanchard2017machine} is an existing outlier detection defense.
\\
\noindent \textbf{Removal:} make minor modifications to the model in order to remove the embedded backdoor. Targets small deviation magnitudes.% Super-fine-tuning (SFT) \cite{sha2022fine} is an existing backdoor removal defense.

Some existing works combine multiple approaches to create more effective defenses; for example, FLAME \cite{nguyen2022flame} first detects malicious updates using cosine-similarity-based clustering, then limits the impact of selected updates by clipping their norms to the median norm and adding noise. We study FLAME and other existing defenses, and their weaknesses in more detail in Section \ref{sec:adaptive_attacks}.

An important consideration is the location in the FL pipeline where a defense operates: pre-aggregation, in-aggregation, or post-aggregation \cite{nguyen2024backdoor}, with detection defenses often operating pre-\\aggregation, robust-aggregation defenses operating in-aggregation, and removal defenses operating post-aggregation.

\subsubsection{Clipped-Super-Fine-Tuning} \label{subsubsec:csft}

\begin{figure}[t!]
    \centering
    \includegraphics[width=0.95\columnwidth]{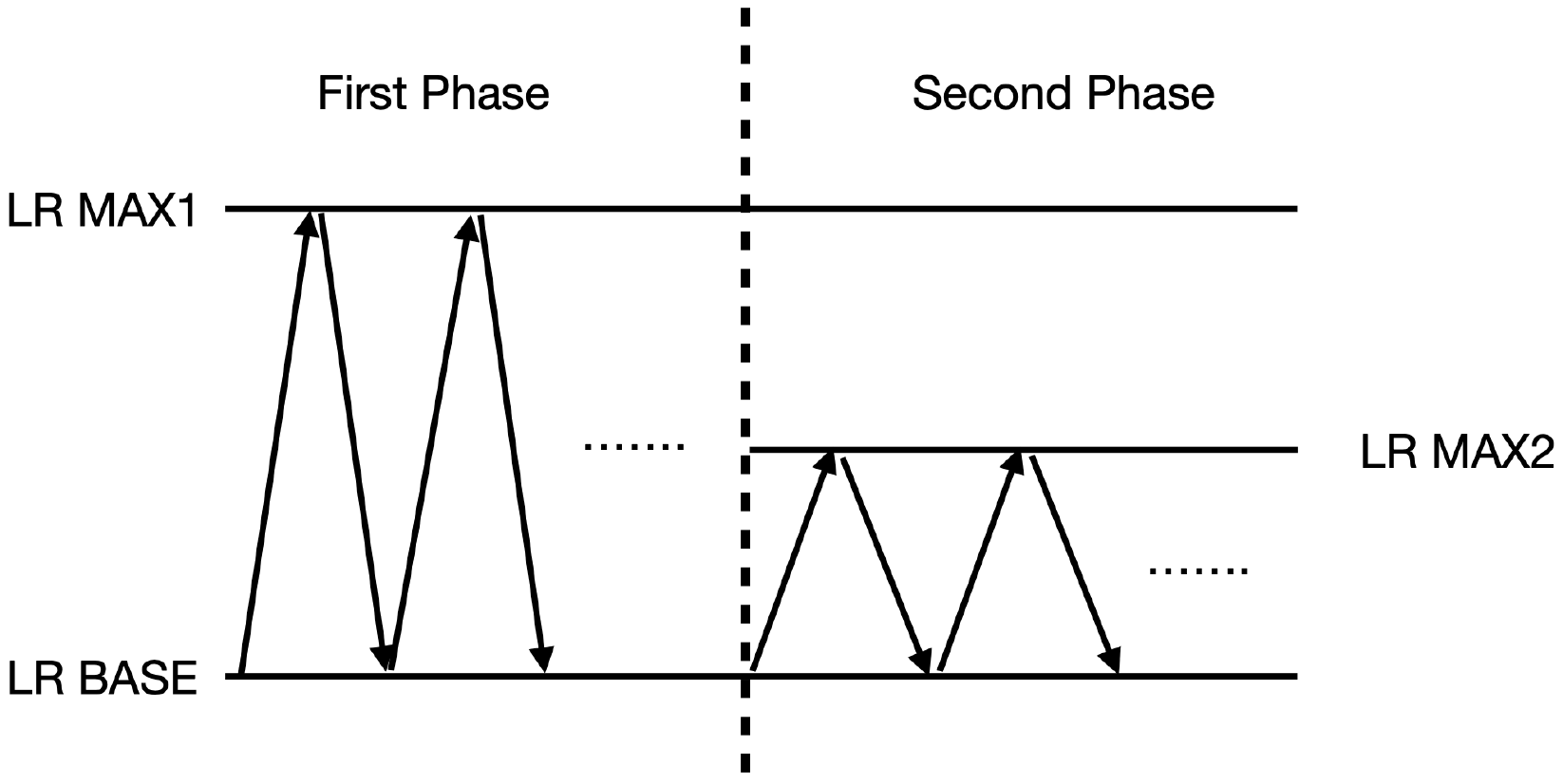}
    \caption{Learning rate scheduling of super-fine-tuning, taken from the original paper \cite{sha2022fine}.}
    \label{fig:super_fine_tuning}
\end{figure}

We adapt \textbf{Super-Fine-Tuning (SFT)} \cite{sha2022fine}, a removal defense designed for centralized training, to the federated setting. SFT removes backdoors by fine-tuning the global model on a small benign dataset using a specialized learning rate schedule, as illustrated in Figure \ref{fig:super_fine_tuning}. However, we find that vanilla SFT performs poorly in FL; the high learning rate required during the first phase often leads to gradient explosion, destroying the model weights. This behavior suggests either that the SFT algorithm is highly sensitive to hyperparameters, or that the loss landscape of models trained in a federated setting is fundamentally different from that of models trained centrally and does not respond to SFT in the same way. 

\begin{figure}[t!]
    \centering
    \includegraphics[width=0.95\columnwidth]{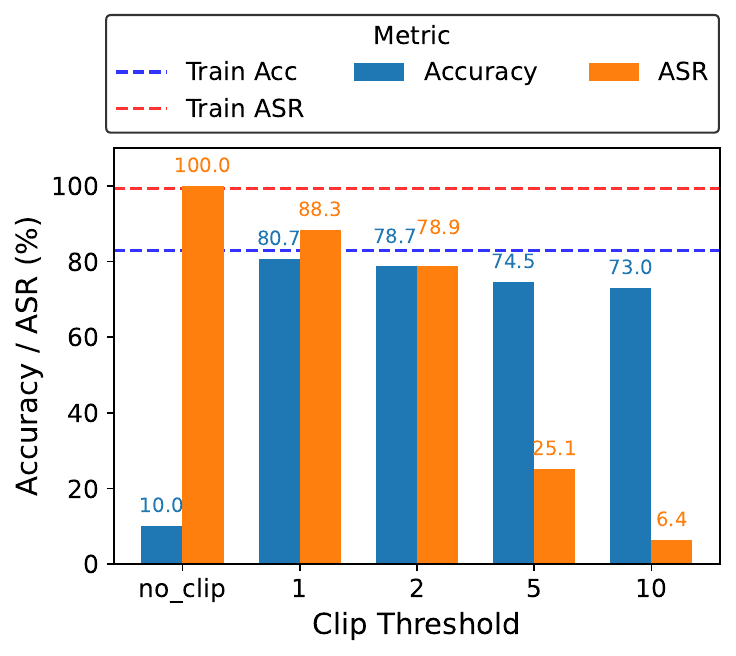}
    \caption{Federated models backdoored with $m=1$ malicious clients, fine-tuned with 2000 samples over different gradient clipping thresholds.}
    \label{fig:fine_tune_clip_thresholds_easy}
\end{figure}

To mitigate this instability, we introduce \textbf{Clipped-Super-Fine-Tuning (CSFT)}, which incorporates gradient clipping into the SFT process. With an appropriately tuned clipping threshold, CSFT effectively removes "weakly inserted" backdoors (small devations), as demonstrated in Figure \ref{fig:fine_tune_clip_thresholds_easy}, where CSFT successfully removes a non-adaptive Badnet \cite{gu2019badnets} backdoor implanted by a single malicious attacker out of 20 in a ResNet18 \cite{resnet} on the CIFAR-10 dataset \cite{cifar10}.
Notably, we find that the optimal clipping threshold depends primarily on the number of fine-tuning samples, significantly reducing the need for exhaustive hyper-parameter searches.

\subsection{Limitations and Assumptions in Federated Learning}\label{subsec:limitations}

The distributed nature of federated learning makes defending against backdoors particularly challenging, as the aggregator has limited visibility into individual clients' behavior. Additionally, thanks to their full control over malicious clients, attackers can deploy more potent attack methods that extend beyond simply adding trigger patterns to the data and flipping labels.

\paragraph{Attacker limitations and assumptions.} A common assumption in federated learning security is that the majority of clients are benign (also called trustworthy) \cite{wang2022flare}; therefore, an attacker can only control a malicious minority of clients. Formally, a subset of $m \leq \lfloor \frac{n-1}{2}\rfloor $ malicious clients submit model updates to the aggregator. Krum \cite{blanchard2017machine}, one of the defenses we study in this work, assumes that $2m+2 < n$, meaning $m < \frac{n}{2} - 1$. Since we run our experiments with $n=20$, we use a tighter bound of $m \leq 8$ across our experiments rather than the $m \leq 9$ offered by the malicious minority assumption.
Another limitation for the attacker is the requirement to preserve benign accuracy. Otherwise, the attack is easily detectable, and the model is not deployed, meaning the attacker cannot exploit it. 

\paragraph{Defender limitations and assumptions.} 
We assume the aggregator has a small, benign dataset drawn from $D$ for fine-tuning, as in MASA \cite{xu2025identify}. In practice, the aggregator could be a trusted client using its own data for fine-tuning, or the data could be sourced from public proxy data or a small set of opt-in user data. To maintain the necessity of the federated framework, this dataset must be a fraction of the total training data (e.g., 1\%-4\%). If the aggregator held a larger portion, a centralized model could be trained locally, rendering federated learning obsolete. In our experiments, we strictly limit the fine-tuning set to be smaller than the average client's local dataset. For instance, our 2000-sample (4\%) set is below the average per-client sample count of 2,400, effectively treating the aggregator as a single trusted client. We study the effect of the number of fine-tuning samples in Section \ref{subsec:csft_hyper}. Furthermore, we acknowledge that while backdoors can always be removed through extreme weight transformations (e.g., random re-initialization), such measures are trivial and provide no utility. A viable defense must therefore preserve benign accuracy while preventing backdoors.

\section{Motivation \& Threat Model}\label{sec:motivation}
Achieving trustworthy and secure FL requires a principled approach grounded in worst-case assumptions. Mirroring the design of cryptographic schemes, we assume a nigh-omniscient attacker to establish a blueprint for security; a defense robust against this adversary will inherently withstand weaker, more realistic threats.

We model an adaptive adversary that controls a subset of $m$ colluding malicious clients, aiming to manipulate the FL training process via malicious updates. To evaluate defenses under worst-case conditions, we grant this adversary full information: exact knowledge of the aggregation algorithm, its specific parameters, and the benign updates submitted by all other clients in each epoch. In Section \ref{sec:def_results}, we demonstrate that our principled defense approach can reliably defeat this full-information, adaptive adversary.

\section{\defensename: Toward a Principled Theory}\label{sec:principle}

\input{tikz_figures/delta_gap}

Each backdoor attack can be categorized according to a set of parameters, including its impact on the model weights, the fraction of malicious clients, its duration, etc.
We consider the impact on model weights to be particularly important in the design of our theory. 
Let the expected deviation from benign updates $\delta$ be
\begin{equation}
\begin{split}
    \delta = \Vert \mathbb{E}[\frac{1}{n-m}\sum_{i=1}^{n-m}T(w_j, d_i)] \,- \qquad \,\\
    \mathbb{E}[\frac{1}{m}\sum_{i=n-m+1}^{n}T(w_j, d_i \cup b_i)]\Vert_2\
\end{split}
\end{equation}
for any given federated epoch $j$, where we have $n$ clients and each client has a training dataset $d_i \sim D$, and we assume there are $m < \frac{n}{2}$ malicious clients with datasets $b_i \sim B$ drawn from the distribution of backdoored samples $B$. 
$\delta$ represents the expected distance from benign updates (or strength) of a given backdoor attack update, with smaller $l_2$-norm deviations having a small $\delta$, and larger $l_2$-norm deviations having a larger $\delta$.

Our principled combined defense is premised on the observation that the success of existing defenses depends on the attack parameter $\delta$.
We define two fundamental types of defenses based on their success relative to $\delta$:

\paragraph{\textbf{Type 1 - Robust Aggregation or Detection:}}
The goal of the aggregator is to discard malicious updates and update the global model only with benign updates, or at least limit the impact of malicious updates on the aggregation process. To do so, the aggregator attempts to identify outliers (which are likely malicious updates, since $m < \frac{n}{2}$) from the client updates $\{T(w_j, d_i)\}_{i=1}^{n-m} \cup \{T(w_j, d_i \cup b_i)\}_{i=n-m+1}^{n}$. It then removes those outliers, e.g., Krum and Multi-Metrics, or bounds their impact, e.g., Median-of-Means and norm bounding.

\paragraph{\textbf{Type 2 - Removal:}}
Assume the aggregator has a small, benign dataset $f \sim D$ and fine-tunes the global model $w$ after aggregation as follows:
\begin{equation}
    w_{\tau} = (1-\gamma)\, w + \gamma \, T(w, f)
\end{equation}
for some $0 \leq \gamma \leq 1$.

Using the distance $\delta$, we can make the following propositions about existing defenses:

\begin{proposition}\label{prop:large_defense_1}
    The larger $\delta$, the more robust Type 1 defenses are.
\end{proposition}
Proposition \ref{prop:large_defense_1} follows from two observations: the larger $\delta$ is, the further in the $l_2$-norm the malicious updates are from the benign updates. Since $m < \frac{n}{2}$, the benign updates outnumber the malicious updates. As such, the further the malicious updates are from the benign updates, the more likely they are to be detected as outliers by outlier detection defenses.
Therefore, a larger $\delta$ results in a lower ASR relative to the equivalent undefended global model.
Robust aggregation defenses impose a limit on the influence any one update can have on the aggregation process. Since the defense's limit naturally bounds $\delta$, the larger the $\delta$, the better the defense relative to the equivalent undefended global model.
Thus, it follows that a large $\delta$ improves the effectiveness of Type 1 defenses, and Corollary \ref{cor:large_defense_1} and Corollary \ref{cor:large_defense_1_attack} also follow. 

\begin{corollary} \label{cor:large_defense_1}
    For all Type 1 defenses, there exists $\delta_1$ such that it will remove the backdoor for any $\delta > \delta_1$.
\end{corollary}

\begin{corollary} \label{cor:large_defense_1_attack}
    For all Type 1 defenses that admit learning (i.e., at least one update), there exists $\delta'_1$ such that a backdoor update with $\delta < \delta'_1$ will pass the defense.
\end{corollary}

\begin{proposition}\label{prop:small_defense_2}
    Let $\varepsilon \in \mathbb{R}^k$ be a matrix of small random values, i.e $\Vert\varepsilon\Vert_2 \ll \Vert w\Vert_2$.
    The smaller $\delta$, the more the global model approaches $w_{\tau} = T(\cdot, f) + \varepsilon$ under a Type 2 defense, i.e., a model trained on a dataset drawn only from $D$.
\end{proposition}

Proposition \ref{prop:small_defense_2} follows from the iterative learning process of neural networks that slowly converge after many epochs. Intuitively, a Type 2 defense can remove the backdoor because $f$ is drawn from the benign distribution $D$, thereby shifting the global model weights towards those of a model trained only on $D$.
The larger the change introduced by the backdoor, the more backpropagation steps are needed to remove it; consequently, the defender faces a higher risk of benign accuracy degradation. Hence, Type 2 defenses are most effective against weakly inserted backdoors, where $\delta$ is small enough to be removed without overwriting the model with one trained solely on the fine-tuning data.

\begin{corollary} \label{cor:small_defense_2}
    For all Type 2 defenses, there exists a $\delta_2$ such that it will remove the backdoor for any $\delta < \delta_2$.
\end{corollary}

\begin{corollary} \label{cor:small_defense_2_attack}
    For all Type 2 defenses, there exists a $\delta'_2$ such that a backdoor update with $\delta > \delta'_2$ will not be removed.
\end{corollary}

Given either only a Type 1 or Type 2 defense, it follows from Corollary \ref{cor:large_defense_1_attack} and Corollary \ref{cor:small_defense_2_attack} that an attacker should be able to succeed. We empirically demonstrate such attacks against existing defenses in Section~\ref{sec:adaptive_attacks} using optimization, scaling, or interpolation to adapt to the protection objectives.

When combining a Type 1 defense and a Type 2 defense, it follows from Corollary \ref{cor:large_defense_1} and Corollary \ref{cor:small_defense_2} that an attacker can only succeed for $\delta \in [\delta_2, \delta_1]$. Figure \ref{fig:security_gap} illustrates this defensive spectrum; while each category effectively secures one end of the magnitude range, a gap remains where neither succeeds.

\noindent \newline \textbf{Research Question 1:} \label{rq:1} \textit{Can we reliably identify a successful principled attack given any single-typed or non-principled combined defense?} \\
\noindent \newline \textbf{Research Question 2:} \label{rq:2} \textit{When combining Type 1 and 2 defenses, can $\delta_1 - \delta_2$ be non-positive, meaning that even worst-case adaptive adversaries cannot succeed?} \\

We hypothesize that with the correct defenses, the gap in our second research question can be eliminated entirely.

\begin{conjecture}\label{conj:1}
    There exists a Type 1 defense and a Type 2 defense such that $\delta_2 \geq \delta_1$.
\end{conjecture}

If Conjecture \ref{conj:1} holds, it implies the existence of a principled defense that can defend against any backdoor for any attack deviation $\delta$. 

\input{tikz_figures/attack_intuition}
\section{Exploiting the Gap in Existing Defenses} \label{sec:adaptive_attacks}
Our goal in this section is to provide empirical evidence to support Corollary \ref{cor:large_defense_1_attack} and Corollary \ref{cor:small_defense_2_attack} by showing that for all existing defenses we evaluate, we can define an adaptive attack that easily breaks the defense. To evaluate the limits of current defenses, we develop a suite of adaptive attacks designed to exploit the fundamental weaknesses of each defense category. By leveraging the full information capabilities of our adaptive adversary, these attacks bypass singular and non-principled security mechanisms. In the following, we outline the general principles for circumventing each category, then detail the specific implementation for each defense.
We display simplified representations of the attack principles for each defense category in Figure \ref{fig:adaptive_attack_principles}.

\subsection{Adaptive Attack Principles} \label{subsec:attack_principle}
\paragraph{Robust-Aggregation Defenses.}
Robust aggregation mechanisms aim to bound the impact of any individual update on the global model. Our strategy exploits this by ensuring the malicious updates reach the maximum allowable impact bound, as shown in Figure \ref{subfig:robust_aggregation}. We achieve this by scaling the magnitude of the adversarial updates relative to the mean, either through direct weight manipulation or by increasing the local learning rate of malicious clients. 

\paragraph{Outlier Detection Defenses.}
Outlier detection defenses use metrics such as $l_2$-distance, $l_1$-distance, or cosine similarity to isolate and discard malicious updates. We formulate our attacks as a multi-objective optimization problem, maximizing the backdoor signal while minimizing the defense's detection metric. Where possible, we solve this optimization directly using interpolation, as shown in Figure \ref{subfig:outlier_detection}; otherwise, we employ gradient-based techniques by incorporating the defense's scoring function into the malicious clients' loss objective, or a surrogate scoring function if the defense's scoring function is not differentiable.

\paragraph{Removal Defenses.}
Removal and post-processing defenses are constrained by the need to preserve the model's benign accuracy. If the defense applies a transformation that is too significant to the model weights, it risks destroying the utility of the global model. We exploit this constraint by scaling the adversarial updates throughout training, ensuring that the modification required to remove the backdoor exceeds the threshold of safe removal for the defense, as shown in Figure \ref{subfig:removal_defenses}.

\paragraph{Non-Principled Combined Defenses.} 
Several works have attempted to construct robust defenses by assembling multiple defense mechanisms into a single pipeline, such as FLAME \cite{nguyen2022flame}, Fine-tune \& Prune \cite{wu2020mitigating}, and FoolsGold \& Krum \cite{fung2018mitigating}. While these multi-layered approaches often outperform individual defenses, the lack of a grounding principle behind their combination leaves them vulnerable to adaptive adversaries. For example, fine-tuning and pruning are both removal-based, which prevent small $l_2$-norm attacks but leave them vulnerable to large $l_2$-norm attacks; FLAME combines HDBSCAN angle-based clustering, a defense against large angular magnitude attacks, with clipping, a defense against large $l_2$-norm attacks, and DP-noising, a defense against attacks involving a small number of malicious clients. Their combined defense leaves them vulnerable to small $l_2$-norm attacks carried out by many attackers.
\\

Table \ref{tab:defense_attack_summary} contains a summary of the defenses for which we devise adaptive attacks separated by type, with the associated attack strategy and a reference to the relevant equations.

\begin{table}[t]
    \centering
    \caption{Summary of defenses and corresponding attack strategies with the relevant equations. Eq. stands for Equation.}
    \label{tab:defense_attack_summary}
    \setlength{\tabcolsep}{4pt}
    \begin{tabularx}{\columnwidth}{@{}L l L l@{}}
        \toprule
        \textbf{Defense Type} & \textbf{Defense Name} & \textbf{Attack Strategy} & \textbf{Eq.} \\ 
        \midrule
        
        \multirow{2}{=}{Type 1 (Robust Aggregation)} 
        & Norm-bounding \cite{sun2019can} & Scaling & \ref{eq:norm_bounding} \\
        & Median-of-Means \cite{minsker2023efficient} & Scaling & \ref{eq:mom} \\
        \midrule
        
        \multirow{6}{=}{Type 1 (Outlier Detection)} 
        & Krum \cite{blanchard2017machine} & Interpolation & \ref{eq:krum} \\
        & Multi-Metrics \cite{huang2023multi} & Optimization & \ref{eq:multi_metrics} \\
        & MARS \cite{wan2025mars} & Optimization & \ref{eq:mars_adapt} \\
        & MASA \cite{xu2025identify} & Interpolation & \ref{eq:krum} \\
        & BART-FL \cite{mia2025bart} & Optimization & \ref{eq:bart_fl} \\
        & AlignIns \cite{xu2025detecting} & Optimization & \ref{eq:align_in_1}, \ref{eq:align_in_2} \\
        \midrule
        
        Type 2 & CSFT (This work) & Scaling & \ref{eq:mom} \\
        \midrule
        
        \multirow{3}{=}{Non-Principled Combined} 
        & FoolsGold \& Krum \cite{fung2018mitigating} & Interpolation & \ref{eq:krum} \\
        & Fine-tune \& Prune \cite{wu2020mitigating} & Scaling & \ref{eq:norm_bounding} or \ref{eq:mom} \\
        & FLAME \cite{nguyen2022flame} & Optimization \& Interpolation & \ref{eq:flame} \\
        
        \bottomrule
    \end{tabularx}
\end{table}

For each category, we iterate through existing defenses in order of recency and present either a scaling formula, an interpolation formula, or an optimization objective that breaks said defense. We present attacks against robust-aggregation defenses in Section \ref{subsec:rob_agg}, outlier detection defenses in Section \ref{subsec:outlier_det}, removal defenses in Section \ref{subsec:removal_def}, and non-principled combined defenses in Section \ref{subsec:bad_comb_def}. We also include empirical results demonstrating the success of our adaptive attacks in Section \ref{subsec:motivating_results}.

\subsection{Motivating Results: Non-Principled Defenses are Exploitable} \label{subsec:motivating_results}

We demonstrate that existing individual backdoor defenses are insufficient under our defined worst-case threat model. Table \ref{tab:motivating_results} summarizes the vulnerability of \textbf{seven} state-of-the-art or popular defenses \cite{nguyen2022flame, blanchard2017machine, minsker2023efficient, sun2019can, sha2022fine} against our adaptive, optimized attacks on a ResNet18 \cite{resnet} model trained on the CIFAR-10 \cite{cifar10} dataset. 

We do not empirically evaluate six of the 13 adaptive attacks we design for the following reasons. FoolsGold \& Krum \cite{fung2018mitigating} and Fine-tune \& Prune \cite{wu2020mitigating}: we already break one of the components of each of these non-principled combined defenses in a way that the other component cannot cope with; therefore, we expect these two defenses to perform as well as their broken components, which we already empirically evaluate. MARS \cite{wan2025mars}, MASA \cite{xu2025identify}, BART-FL \cite{mia2025bart}, and AlignIns \cite{xu2025detecting}: we leave the experimental effort to future work as we already empirically evaluate two outlier detection defenses (Krum and Multi-Metrics). These four defenses reuse similar fundamental principles, so we expect the principled adaptive attacks we design against them to be as effective as those against Krum and Multi-Metrics.

We break most defenses with as few as 1 or 2 malicious clients out of 20. While some defenses show higher initial resistance, specifically, Multi-Metrics requires $m=4$ to break, they all eventually fail as the number of malicious clients increases, reaching near 100\% ASR in the worst-case scenario where $m=8$, while maintaining high benign accuracy. These failures stem from a fundamental vulnerability in current defense approaches. Most existing methods fall into a single category, such as robust aggregation, outlier detection, or removal, that targets only small or large attack magnitudes, leaving them vulnerable to the other. Furthermore, combined defenses often merge sub-defenses of the same type, leaving a vulnerable gap that a full-information adaptive adversary can exploit.

\input{tables/lowest_m_table_badnet_train}

\subsection{Attacking Robust-Aggregation} \label{subsec:rob_agg}

\paragraph{Norm-bounding (2019).}
Norm bounding, also called norm clipping \cite{sun2019can}, is an $l_2$-norm-based algorithm that defends against large magnitude attacks. In practice, norm-bounding can be efficiently implemented in cryptography using zero-knowledge proofs, making it a desirable defense for secure aggregation protocols \cite{lycklama2023rofl}.
Update norms are clipped to a predefined threshold $\eta$ to limit the impact of any single update during aggregation. 
Our adaptive adversary exploits this clipping by directly scaling each malicious update $u_m$ to this predefined threshold, as described in Equation \ref{eq:norm_bounding}.
To defend against such an attack, a norm-bounding defender would need to use a very small norm threshold, at the cost of a greatly reduced model accuracy. We find that for any functional threshold, a single adaptive malicious client is sufficient to break norm-bounding, as seen in Table \ref{tab:motivating_results}.
\begin{equation} \label{eq:norm_bounding}
    u_m' = \eta \frac{u_m}{\; \,\Vert u_m\Vert_2}
\end{equation}

\paragraph{Median-of-Means (2023).}
Median of Means (MoM) \cite{minsker2023efficient} is a provably robust aggregation algorithm that aims to minimize the impact of any one update on the aggregation regardless of its magnitude. 
For each parameter layer in the neural network model, MoM randomly partitions client updates into groups. 
Then it computes the element-wise mean for each group as its representative. 
Finally, it takes the element-wise median of the representatives as the update for that layer. 
In the MoM algorithm, outlier updates move their groups away from the median, reducing the likelihood that they are selected. 
To improve the robustness of the MoM estimation, we can repeat the MoM algorithm $k$ times and average the medians across the $k$ runs. 
We refer to this method as the Robust Median of Means (Robust MoM, shortened to Rob MoM). 
In our experiments, we set $k=10$ as we did not find further meaningful improvements in robustness with higher values.

Despite MoM provably limiting the impact of malicious updates on the aggregation, our adaptive attacker can still skew the median in the adversarial direction. 
Scaling up the updates of malicious clients following Equation \ref{eq:mom} causes their updates to overpower the others in their group, affecting the representative. 
The change incurred by malicious updates to their group's representative shifts the median parameter selected in the adversarial direction for every model parameter. 
While the change in the median is minor for any individual parameter, the cumulative effect across the high-dimensional weight space results in a significant $l_2$-norm shift in the adversarial direction. 
We find that while the defense is successful when there is a single malicious client among 20, it breaks when we introduce a second malicious client, as shown in Table \ref{tab:motivating_results}. We also did not find any meaningful differences between MoM and Rob-MoM against our adaptive attack.

\begin{equation} \label{eq:mom}
    u_m' = \gamma \cdot u_m, \quad \text{where} \quad \gamma \gg \max_{i \in \{1, 2, \dots, n-m\}} \Vert u_{b_i}\Vert_2
\end{equation} 

\subsection{Evading Outlier Detection} \label{subsec:outlier_det}

\paragraph{Krum (2017).}
Krum \cite{blanchard2017machine} is a Byzantine-resilient detection method that selects the most representative update based on its Krum-score: the sum of the $l_2$ distances to its $n-m-2$ closest neighbors. The update with the lowest score is selected as the update to the global model for that epoch. First, if the attacker controls more than one malicious client, they can overwrite the malicious updates with the mean malicious update. This process eliminates $m-1$ terms from the sum when computing the Krum-score of the malicious updates. Second, suppose the mean malicious update is still not selected after this process. We can interpolate between the mean of the malicious updates and the mean of the benign updates to find the point on the line between the two that is closest to the mean of the malicious updates, while selected by Krum.
We solve this directly using bisection with a maximal depth of 10. Using its full information, the adaptive adversary iteratively simulates the Krum aggregation process to perform an informed search along the vector toward the benign mean. This optimization process ensures the malicious update is selected while being positioned as close as possible to the target weights. Let $w_b$ be the mean benign update and $w_m$ be the mean malicious update. Given the maximal distance $\alpha \in [0, 1]$ to the mean benign update on the line such that Krum still selects the update, found by bisection, we can compute the final malicious update as :
\begin{equation} \label{eq:krum}
    w_m' = \alpha w_m + (1-\alpha)w_b
\end{equation}
As shown in Table \ref{tab:motivating_results}, our adaptive attack is successful against Krum with a single malicious client.

\paragraph{Multi-Metric (2023).}
Multi-Metric \cite{huang2023multi} computes a multi-metric feature vector composed of the pairwise $l_1$-distance, $l_2$-distance, and cosine similarity between each pair of updated models. It then applies a whitening transformation to weight the metrics into a unified score dynamically. Then it sorts the updated models by score and selects the 30\% with the lowest scores. Finally, it computes a weighted average of the selected updates, where each client's weight is proportional to its self-reported training sample count relative to the total training sample count across all selected participants.

The Multi-Metric score computation is entirely differentiable, rendering it vulnerable to gradient-based adaptive attacks. Because the defense relies on relative pairwise distances across the entire pool of updates, the malicious updates directly influence the scores assigned to the benign clients. We exploit this dynamic by formulating an optimization objective that not only minimizes the malicious clients' scores but simultaneously maximizes the benign clients' scores. We incorporate this into the malicious loss function as follows:
\begin{equation} \label{eq:multi_metrics}
\begin{split}
\min_{w} \lambda \mathbb{E}_{(x,y)\in(d \cup b)}[L(x, y)] & + \frac{1}{|W_m|} \sum_{w_i \in W_m} M(w_i, W) \\
    & - \frac{1}{|W_b|} \sum_{w_b \in W_b} M(w_b, W)
\end{split}
\end{equation}
where $\lambda$ is a scaling hyperparameter, $w$ represents the malicious model weights being optimized, $W_m$ and $W_b$ are the sets of malicious and benign model weights respectively, $W = W_m \cup W_b$, and $M(\cdot, W)$ is the differentiable Multi-Metric scoring function.
Doing so is sufficient to break Multi-Metric with four malicious clients, as shown in Table \ref{tab:motivating_results}. 
We also identify a critical integrity flaw in Multi-Metric's design: it relies on self-reported metadata for aggregation weighting. By claiming an artificially large sample size, a selected malicious client can perform a model replacement attack \cite{bagdasaryan2020backdoor}, effectively overwriting the global model weights at no additional cost to the attacker.

\paragraph{MARS (2025).} MARS \cite{wan2025mars} is an outlier detection defense that identifies backdoored updates by evaluating the Backdoor Energy (BE) of individual neurons. To amplify this malicious signal, the aggregator extracts the highest BE values from each layer of the local models to form a one-dimensional Concentrated Backdoor Energy (CBE) vector. The aggregator then separates these vectors into two clusters using K-Means and the 1-Wasserstein distance to capture the overall probability distribution of the energies, regardless of their specific neuron ordering. Finally, the defense aggregates the trusted models by selecting the cluster with the smaller center norm, or defaults to the majority cluster (MARS*) to resist adaptive regularization attacks.

Our adaptive adversary can circumvent MARS by distributing the backdoor signal across a wider set of model parameters, thus flattening the Backdoor Energy (BE) and evading the top-$\kappa\%$ extraction filter. Furthermore, by explicitly minimizing the 1-Wasserstein distance between the malicious and benign CBE distributions (Eq. \ref{eq:mars_adapt}), the attacker can ensure the backdoored update is indistinguishable from the majority cluster, compelling the aggregator to include it in the global model.
\begin{equation} \label{eq:mars_adapt}
    \min_{w} \mathbb{E}_{(x,y)\in(d\cup b)}[L(x, y)] + \frac{1}{n-m} \sum_{i=1}^{n-m} Wass(CBE(w), CBE(w_b^i))
\end{equation}
Here, $Wass$ denotes the 1-Wasserstein distance, $CBE$ refers to the function that returns the Concentrated Backdoor Energy vector, and $w_b^i$ represents the $i$-th benign model.

\paragraph{MASA (2025).} MASA \cite{xu2025identify} is an outlier detection defense that identifies backdoored models by exploiting their distinct behavior during machine unlearning. Since malicious parameters are largely inactive on benign data, backdoored models exhibit significantly higher empirical losses when the aggregator attempts to unlearn the main task using gradient ascent on a benign proxy dataset. To enhance detection robustness, MASA addresses non-IID divergence by fusing local updates with the global average, accumulating the individual unlearning loss for each model, and discarding any update whose Median Deviation Score (MDS) exceeds a predefined threshold.

Analogous to the adaptive attacks against Krum described in Section \ref{subsec:outlier_det} (Equation \ref{eq:krum}), an informed adversary can bypass MASA using a simulation-guided bisection attack. By using their local benign data as a surrogate for the aggregator's private proxy dataset, the attacker can replicate the unlearning trajectory. Through a binary search between the target malicious update and the benign center, as defined in Equation \ref{eq:krum}, the attacker can identify the maximum scaling factor that embeds the backdoor while ensuring the simulated Median Deviation Score (MDS) remains below the rejection threshold.

\paragraph{BART-FL (2025).} BART-FL \cite{mia2025bart} is an outlier detection defense that filters malicious updates using a combination of dimensionality reduction and statistical voting. The aggregator first applies Principal Component Analysis (PCA) to reduce the dimensionality of received updates and then computes pairwise cosine similarities to quantify their directional alignment. These similarity scores serve as features for K-means clustering, which partitions the clients into two groups. To reliably identify the benign cluster, the defense employs a multi-metric statistical voting mechanism based on point-level mean similarity, cluster-level mean similarity, and Median Absolute Deviation (MAD), ultimately aggregating only the updates from the winning cluster.

The outlier detection pipeline of BART-FL, which relies on PCA projections and K-means clustering, can be modeled as a differentiable constraint. Since a full-information adaptive adversary observes the benign updates, they can locally simulate the aggregator's dimensionality reduction. By optimizing malicious updates to minimize the distance to the benign cluster while maintaining sufficient intra-update variance, the attacker can ensure that their updates are included in the selected cluster and that the backdoor signal is preserved. We describe the adversary models' loss function in Equation \ref{eq:bart_fl}. Let $P(\cdot)$ be the aggregator's PCA projection function, $u_b$ be the center of the benign updates, and $W_{-j}$ be the set of malicious clients excluding malicious client $w_j$. 
\begin{equation} \label{eq:bart_fl}
\begin{split}
\min_{w_j} \mathbb{E}_{(x,y)\in(d \cup b)}[L(x, y)] &+ \lambda_1 ||P(u_m) - P(u_b)||_2 \\
&- \lambda_2 \frac{1}{|W_{-j}|} \sum_{w_i \in W_{-j}} ||w_j - w_i||_2
\end{split}
\end{equation}

\paragraph{AlignIns (2025).} AlignIns \cite{xu2025detecting} is an outlier detection defense that identifies backdoored updates by analyzing their directional patterns at two levels of granularity. It first evaluates the overall directional deviation of each local update from the global model ($w_g$) using Temporal Direction Alignment (TDA). Subsequently, it performs a fine-grained Masked Principal Sign Alignment (MPSA) to assess how well the signs of an update's most critical (top-$k$-largest) parameters align with the majority sign across all clients. Finally, AlignIns applies a median-based Z-score to filter out anomalous updates and clips the magnitudes of the remaining updates to ensure robust aggregation.

AlignIns targets directional patterns through TDA and MPSA. Our adaptive attacker can defeat MPSA by isolating the top-$k$ parameters and optimizing them on benign data to mimic benign sign alignments, as described in Equation \ref{eq:align_in_1}. For the remaining parameters, the attacker can embed the backdoor signal while adhering to a directional penalty and norm-clipping as described in Equation \ref{eq:align_in_2}, thereby ensuring the update passes both the TDA filter and the MPSA check.

Let $M_k \in \{0, 1\}^{|w|}$ be a binary mask where $1$ indicates a top-$k$ parameter and $0$ otherwise, and $U_b$ is the set of benign updates. The attacker splits the training into two constrained objectives.\\
For the top-$k$ parameters ($w_{top} = M_k \odot w$):
\begin{equation} \label{eq:align_in_1}
    \min_{w_{top}} \mathbb{E}_{(x,y)\in d}[L(x,y)]
\end{equation}
For the remaining parameters ($w_{rest} = (1 - M_k) \odot w$):
\begin{equation}\label{eq:align_in_2}
\begin{gathered}
  \min_{w_{rest}} \mathbb{E}_{(x,y)\in(d \cup b)}[L(x,y)] - \lambda \cdot \text{TDA}(u_m, w_g),\\
  \text{subject to } \max(|w_{rest}|) < \min(|w_{top}|), \\
  \text{ and } ||u_m||_2 \le \text{Median}_{u_b \in U_b}(||u_b||_2)
  \end{gathered}
\end{equation}

\subsection{Circumventing Removal Defenses} \label{subsec:removal_def}
\paragraph{CSFT (This work).} Removal-based defenses attempt to purge backdoors post-aggregation or post-training by applying transformations such as fine-tuning or pruning to the global model weights. 
However, the efficacy of these methods is fundamentally constrained by the accuracy-robustness trade-off: a defender must remove the malicious signal without damaging the model's benign performance. 

Our adaptive adversary exploits this constraint by optimizing for backdoor durability. 
By strategically increasing the magnitude of malicious updates during training, following Equation \ref{eq:mom}, the adversary ensures the backdoor signal is deeply embedded within the model's weights. 
This optimization forces a scenario in which any weight modification significant enough to eliminate the backdoor would concurrently significantly degrade the model's utility. 
We demonstrate this principle using CSFT, as described in Section \ref{subsubsec:csft}. 
As shown in Table \ref{tab:motivating_results}, our adaptive attack embeds a sufficiently durable payload to break CSFT with only a single malicious client while maintaining high benign accuracy.

\subsection{Breaking Non-Principled Combined Defenses} \label{subsec:bad_comb_def}
\paragraph{FoolsGold \& Krum (2018).}
FoolsGold \& Krum is designed to use FoolsGold to mitigate sybil attacks (multiple malicious clients) and Krum to handle single-attacker scenarios, providing, in theory, coverage against attacks with any number of malicious clients.
However, we show in Section \ref{subsec:rob_agg} (Equation \ref{eq:krum}) how to defeat Krum, and in Table \ref{tab:motivating_results}, we demonstrate that we can do it with only a single malicious client. 
Therefore, combining FoolsGold with Krum provides no additional benefit over using Krum alone, and the combined defense is broken by construction.

\paragraph{Fine-tune \& Prune (2020).}
Fine-pruning and fine-tuning are both post-aggregation defense strategies primarily effective against small-magnitude updates. Consequently, our adaptive strategy for circumventing removal defenses (Section \ref{subsec:removal_def}) following Equation \ref{eq:norm_bounding} or Equation \ref{eq:mom} remains effective against this combination. This highlights that observing the principles for combination is key to designing robust combined defenses.

\paragraph{FLAME (2022).} FLAME \cite{nguyen2022flame} integrates HDBSCAN angle-based clustering, median-based clipping, and Differentially-Private (DP) noising. While these components effectively mitigate large angular deviations and high-$l_2$-norm updates, they remain collectively blind to subtle attacks carried out by a larger number of malicious clients. We model this vulnerability as a differentiable optimization problem where the adversary must simultaneously: (1) minimize cosine similarity to the selected cluster to evade HDBSCAN, (2) maximize $l_2$-norm updates to counteract median-based clipping (similar to our attack in Section \ref{subsec:rob_agg}), and (3) reinforce the backdoor signal to withstand DP-noising. During training, we incorporate these objectives into the malicious clients' loss function:
\begin{equation} \label{eq:flame}
\begin{split}
    \min_{w} \mathbb{E}_{(x,y)\in(d\cup b)}[L(x, y)] &+ \frac{\alpha}{|C|}\sum_{w_c \in C}(1 - \text{cos}(w, w_c)) \\
    &+ \beta ~ \text{ReLU}(l - \Vert u \Vert_2)^2
\end{split}
\end{equation}
where $w$ represents the malicious model weights, $u$ its update, $C$ the set of benign weights identified by the clustering algorithm, and $\alpha$ and $\beta$ are hyperparameters. As shown in Table \ref{tab:motivating_results}, our adaptive attack breaks FLAME with only two malicious clients out of twenty while preserving benign accuracy. For the MNIST dataset \cite{mnist} in Section \ref{subsec:mnist}, we find it helpful to interpolate between the mean malicious update and the $l_2$-projected mean of the closest benign neighbors when the malicious updates are not selected.

\section{\defenseshortname: A Principled Combined Defense} \label{sec:def_results}
We mitigate the gaps left by existing non-principled defenses with our principled defense approach: \defensename~ (\defenseshortname). \defenseshortname~ combines defenses based on their categories, such that no defense's vulnerability is exploitable. While Type 1 defenses are well-represented by robust aggregators (e.g., Median-of-Means \cite{minsker2023efficient}), outlier detection algorithms (e.g., Krum \cite{blanchard2017machine}, Multi-Metrics \cite{huang2023multi}), or hybrid pipelines like FLAME \cite{nguyen2022flame}, there are few Type 2 defenses, and even fewer tailored for the unique constraints of FL. By instantiating our Type 2 defense as CSFT, as described in Section \ref{subsubsec:csft}, we can systematically pair it with the Type 1 defenses evaluated in Table \ref{tab:motivating_results}, namely FLAME \cite{nguyen2022flame}, (Rob) MoM \cite{minsker2023efficient}, Krum \cite{blanchard2017machine}, Multi-Metrics \cite{huang2023multi}, and Norm-bounding \cite{sun2019can}, yielding: \textbf{$\text{HA}_{\text{Flame}}^{\text{CSFT}}$}, \textbf{$\text{HA}_{\text{(Rob) MoM}}^{\text{CSFT}}$}, \textbf{$\text{HA}_{\text{Krum}}^{\text{CSFT}}$}, \textbf{$\text{HA}_{\text{Multi-Metrics}}^{\text{CSFT}}$}, and \textbf{$\text{HA}_{\text{Norm}}^{\text{CSFT}}$}.
The resulting principled combined defenses support our theory. They offer empirical evidence for Corollary \ref{cor:large_defense_1}, Corollary \ref{cor:small_defense_2}, and Conjecture \ref{conj:1}.

% argue why adaptive attacks are adaptive to our principled combined defense
\begin{proposition} \label{prop:ha_adaptive_attack}
    A full-information adaptive attack that is optimal against a Type 1 defense A is also optimal against $\text{HA}_{\text{A}}^{\text{B}}$, for any Type 2 defense B.
\end{proposition}

Proposition \ref{prop:ha_adaptive_attack} follows from Corollary \ref{cor:large_defense_1}: for an adaptive attack to be optimal against a Type 1 defense, it needs to maximize $\delta < \delta_1$ to maximize its impact on the global model; and it follows from Corollary \ref{cor:small_defense_2}: for an adaptive attack to be optimal against a Type 2 defense, it needs to maximize $\delta$ to maximize its impact on the global model such that $\delta > \delta_2$. Consequently, if there exists a maximum reachable $\delta$ such that $\delta_2 < \delta < \delta_1$, an optimal adaptive attack against the Type 1 defense will find it. Hence, if our full-information adaptive attacks are optimal against Type 1 defenses, they must also be optimal against \defenseshortname.

% Introduce the results and the rest of the evaluation sections
We demonstrate in the following sections that \defensename~ is effective at combining compatible defenses across a wide range of scenarios. In Section \ref{subsec:eval_conj_1}, we provide empirical evidence supporting Conjecture \ref{conj:1}. We show that even in the worst-case scenario: our full-information adaptive adversary with $m=8$ out of $n=20$, our \textbf{three} best principled combined defenses: \textbf{$\text{HA}_{\text{Flame}}^{\text{CSFT}}$}, \textbf{$\text{HA}_{\text{Krum}}^{\text{CSFT}}$}, and \textbf{$\text{HA}_{\text{Multi-Metrics}}^{\text{CSFT}}$} are undefeated.
Then, we evaluate \defenseshortname~ against existing popular and state-of-the-art non-adaptive attacks in Section \ref{subsec:existing_attacks}. We demonstrate that they are completely ineffective against \defenseshortname, even in settings that heavily favor the adversary.

To support Conjecture \ref{conj:1} with further empirical evidence, we experiment with a non-IID data distribution of CIFAR-10 in Section \ref{subsec:non_iid} and the MNIST dataset \cite{mnist} in Section \ref{subsec:mnist}. The non-IID data setting is common in the FL security literature, as it represents a more realistic deployment scenario \cite{nguyen2022flame, xu2025detecting, wan2025mars, xu2025identify, huang2023multi}. The MNIST dataset is popular for extending experiments to other datasets \cite{nguyen2022flame, xu2025detecting, wan2025mars}. We focus exclusively on the best combined defenses: $\text{HA}_{\text{Flame}}^{\text{CSFT}}$, $\text{HA}_{\text{Krum}}^{\text{CSFT}}$, and $\text{HA}_{\text{Multi-Metrics}}^{\text{CSFT}}$.

We also present additional empirical studies on the effects of hyperparameters on \defenseshortname, specifically regarding CSFT, to measure ease of use in real-world scenarios. We measure the effect of the size of the fine-tuning dataset and the number of fine-tuning epochs on the performance of $\text{HA}_{\text{Flame}}^{\text{CSFT}}$, $\text{HA}_{\text{Krum}}^{\text{CSFT}}$, and $\text{HA}_{\text{Multi-Metrics}}^{\text{CSFT}}$ on CIFAR-10.

\input{tables/lowest_m_simplified_table}

\subsection{Evaluating \defenseshortname: Empirical Evidence Supporting Conjecture \ref{conj:1}} \label{subsec:eval_conj_1}

We present the evaluation of our combined \defenseshortname~ architectures against our full-information adaptive adversaries on CIFAR-10 in Table \ref{tab:lowest_m_simplified}. The results demonstrate that our principled approach successfully closes the exploitable gap in \textbf{three} configurations ($\text{HA}_{\text{Flame}}^{\text{CSFT}}$, $\text{HA}_{\text{Krum}}^{\text{CSFT}}$, and $\text{HA}_{\text{Multi-Metrics}}^{\text{CSFT}}$), rendering the full-information adaptive attacker entirely unsuccessful even in the worst-case scenario ($m = 8$). In another configuration, the gap is significantly narrowed ($\text{HA}_{\text{Norm}}^{\text{CSFT}}$ now requires $m=4$, up from $m=1$). Conversely, pairing CSFT with MoM or Rob-MoM yields little improvement, indicating that the success of our defense approach ultimately relies on selecting a Type 1 defense restrictive enough to overlap with CSFT's effective $\delta$-range. Ultimately, these findings provide supporting empirical evidence that \textbf{Conjecture \ref{conj:1} holds true}: through our principled \defenseshortname~ defense approach, \textit{\textbf{it is possible to defend against a worst-case, full-information, adaptive adversary with minimal degradation to benign accuracy}}. 

\subsection{Evaluating \defenseshortname~ against Non-Adaptive Attacks}\label{subsec:existing_attacks}

We evaluate the constraint-and-scale attack (C\&S) \cite{bagdasaryan2020backdoor} (large deviation attack), the model-replacement attack (Replacement) \cite{bagdasaryan2020backdoor} (large deviation attack), the Distributed Backdoor Attack (DBA) \cite{xie2019dba} (large deviation attack), and the Neurotoxin attack \cite{zhang2022neurotoxin} (small deviation attack). We present the results in Table \ref{tab:existing_attacks_m8}, where we evaluate them against our three best \defenseshortname~variants in the worst-case scenario: $m=8$ malicious clients out of $n=20$.
We find that none of them are successful, with $\text{HA}_{\text{Multi-Metrics}}^{\text{CSFT}}$ displaying the highest benign accuracies across all attacks. These empirical results further reinforce the defensive capabilities of \defenseshortname.

\input{tables/existing_attacks_m8_table}

\subsection{\defenseshortname~ in the Non-IID Data Setting} \label{subsec:non_iid}
\input{tables/lowest_m_simplified_non_iid}

In the real world, data is unlikely to be distributed independently and identically (IID) across clients. As such, we study whether \defenseshortname's performance is affected by a non-IID data distribution across clients.
We follow previous work \cite{xie2019dba,xu2025identify} and use a Dirichlet distribution with a hyperparameter of 0.5 to assign data samples to the clients. 

We present our results in Table \ref{tab:lowest_m_simplified_non_iid} and demonstrate that even under non-IID conditions: 1) our full-information adaptive adversary succeeds against individual defenses, with the needed number of malicious clients ranging from 1 to 4; 2) all three of our combined defenses are successful against our adaptive attacks regardless of the number of malicious clients, even in the worst-case scenario of $m=8$. 

\subsection{\defenseshortname~ on MNIST: Further Evidence Supporting Conjecture \ref{conj:1}} \label{subsec:mnist}
We evaluate \defenseshortname~further on a different dataset: MNIST \cite{mnist}, using a LeNet \cite{lecun2002gradient} model. We show the lowest number of malicious clients needed to break our best \defenseshortname~variants in Table \ref{tab:lowest_m_simplified_mnist}. We find that HA is successful against our adaptive attacks, mirroring our results on CIFAR-10 in Section \ref{subsec:eval_conj_1}, providing further empirical evidence for Conjecture \ref{conj:1}, except that our adaptive attack against FLAME is less successful. We posit that this change is due to the simplicity of the MNIST classification task and the LeNet architecture's lower parameter count compared to ResNet, which makes it more difficult for the adversary to hide the backdoor.

\input{tables/lowest_m_simplified_mnist}

\subsection{CSFT Hyperparameter Study} \label{subsec:csft_hyper}
We empirically validate our hyperparameters for CSFT and show that it requires no additional tuning when the Type 1 defense we combine with changes. We are mainly concerned with the fine-tuning dataset size and the number of fine-tuning epochs. We show that 4\% of the total training set size, which is less than the data of one client (5\%), works almost best for our best \defenseshortname~ variants: $\text{HA}_{\text{Flame}}^{\text{CSFT}}$, $\text{HA}_{\text{Krum}}^{\text{CSFT}}$, and $\text{HA}_{\text{Multi-Metrics}}^{\text{CSFT}}$. More fine-tuning data yields only marginal improvements. We also show that simply picking the epoch with the highest benign accuracy after fine-tuning for 100 epochs is empirically optimal to maximize benign accuracy while keeping ASR below the 50\% threshold.

\begin{figure*}    
    \centering
    \includegraphics[width=0.99\textwidth]{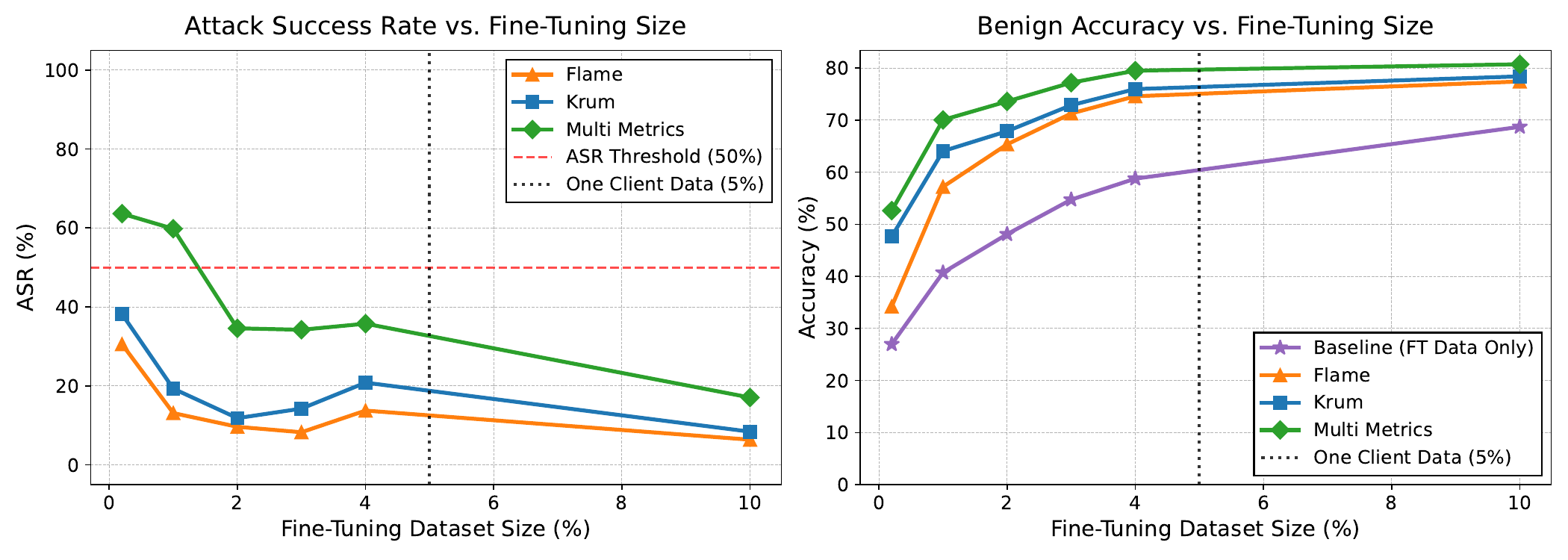}
    \caption{Impact of varying the Fine-Tuning (CSFT) dataset size on Attack Success Rate (left, $m=8$) and Benign Accuracy (right) for $\text{HA}_{\text{Flame}}^{\text{CSFT}}$, $\text{HA}_{\text{Krum}}^{\text{CSFT}}$, and $\text{HA}_{\text{Multi-Metrics}}^{\text{CSFT}}$ on CIFAR-10, alongside a baseline trained centrally solely on the fine-tuning subset, averaged over five seeds. Dataset sizes are expressed as a percentage of the total training data. The vertical dotted line represents the equivalent data held by a single local client (5\%).}
    \label{fig:varying_ft}
\end{figure*}

\subsubsection{Effect of Fine-Tuning Dataset Size}
A concern with defenses that require benign data, such as CSFT or MASA \cite{xu2025identify}, is the amount of data necessary for the defense to function effectively. We show in Figure \ref{fig:varying_ft} that \defenseshortname~'s benign accuracy and defense success rate rapidly converge as fine-tuning data amounts increase, with amounts higher than 4\% of the total dataset size (less than the training set of a single client) providing almost no additional benefits.

\subsubsection{When To Stop Fine-Tuning ?}

By default, we fine-tune with CSFT for 100 epochs across our experiments. However, since the defender cannot measure the ASR in practice to know when to stop fine-tuning, we need to verify that our choice of the number of fine-tuning epochs is robust across defenses. To that end, we compute the safe optimal epoch to stop at, i.e., the epoch that maximizes benign accuracy while the ASR is below 50\%. We compare the safe optimal epoch to the peak epoch, which is the epoch with the highest benign accuracy regardless of ASR. We display the results in Table \ref{tab:optimal_early_stopping}. We find that for all \defenseshortname~ variants and for all numbers of malicious clients, the safe and peak stopping epochs are exactly the same. Hence, CSFT requires no surrogate metric or process to determine when to stop fine-tuning; simply picking the epoch with the highest benign accuracy is empirically optimal.

\input{tables/optimal_early_stopping_table}

\section{Related Work}\label{sec:related_work}
The research community has recently shown a great interest in federated learning backdoors, from attacks
\cite{zhuang2023backdoor,liu2024beyond,nguyen2023iba,xie2019dba,zhang2022neurotoxin,bagdasaryan2020backdoor,gong2022coordinated,zhang2020poisongan,wang2020attack,bhagoji2019analyzing,zhou2021deep,baruch2019little,li20233dfed,zhang2023a3fl,dai2023chameleon,xu2022more,liu2023facilitating,zhao2022defeat,salem2022dynamic,sun2022semi,chang2024fedtrojan,wang2025sba, krauss2024automatic, krauss2023mesas, gu2019badnets} to defenses \cite{nguyen2022flame,xu2025detecting,minsker2023efficient,blanchard2017machine,kumari2023baybfed,fereidooni2023freqfed,lyu2023poisoning,fang2023vulnerability,li2024darkfed,huang2025scope,yu2023g,munoz2019byzantine,shen2016auror,fung2018mitigating,sattler2020byzantine,tolpegin2020data,bagdasaryan2019differential,preuveneers2018chained,nguyen2019diot,zhang2022fldetector,yin2018byzantine,guerraoui2018hidden,ozdayi2021defending,bernstein2018signsgd,rieger2022deepsight,xie2021crfl,sun2021fl,wu2020mitigating,liu2018fine,sun2019can,li2020learning,wu2022federated,cao2019understanding,chen2017distributed,pillutla2022robust}, and surveys \cite{xie2024survey,tariq2024trustworthy,feng2025survey,nguyen2024backdoor,li2025backdoor,wan2024data,szelkag2025adaptive,chen2023investigation}. 

Existing works \cite{zhang2022neurotoxin, bagdasaryan2020backdoor, szelkag2025adaptive, krauss2024automatic, krauss2023mesas, xie2019dba, gu2019badnets} have focused on adversaries with limited capabilities, with at most complete control over the weights they submit to the aggregator, and on other malicious clients, if any exist \cite{li2025backdoor}. Such adversaries help defenders estimate their robustness against an \textit{average} real-world adversary. Such estimates might not apply to a worst-case adversary. Also, they do little for \textit{worst-case} security and for studying the limits of backdoor attacks and defenses, as we do in this work. Several works have raised concerns about the inadequacy or mismatch to reality of existing threat models \cite{krauss2023mesas, krauss2024automatic, szelkag2025adaptive}, underpinning the relevance of our full-information adaptive adversary threat model.

Adaptive attacks \cite{szelkag2025adaptive, krauss2023mesas, krauss2024automatic, shejwalkar2021manipulating} have recently emerged and have been shown to be devastating. Attackers can leverage the disconnecting and reconnecting process to adapt to the defense used by the aggregator \cite{szelkag2025adaptive}, utilize defense metrics to adjust attack parameters \cite{krauss2024automatic}, or adapt through multi-objective optimization \cite{krauss2023mesas}. Adaptive attacks have been shown to break existing state-of-the-art defenses \cite{krauss2023mesas}. For poisoning attacks, a similar threat model to our full-information adaptive adversary was studied, and it defeated all Byzantine-robust aggregation methods evaluated \cite{shejwalkar2021manipulating}. Our work's empirical study extends their evaluation to backdoor attacks, considering also outlier detection, removal, and combined defenses. All current adaptive attacks attack only single-typed defenses. Therefore, according to our theory, if well-formed, they must succeed (as they do).

Few works study defenses against adaptive adversaries. MESAS \cite{krauss2023mesas} considers strong adversaries that adapt via multi-objective optimization; however, it does not use the benign updates that our attacks leverage. 
As a defense, MESAS utilizes a plethora of metrics derived from client updates to identify and discard malicious updates, functioning similarly to Multi-Metrics \cite{huang2023multi}. Unfortunately, MESAS' code is not publicly available; however, if our theory of backdoors is correct, one can defeat MESAS in a similar manner to Multi-Metrics using our full-information adaptive adversary with an optimization-based attack strategy, as we did in Section \ref{subsec:outlier_det}.

\section{Conclusion}\label{sec:conclusion}
In this work, we propose \textbf{\defensename}, a new theory of backdoors in FL based on the expected update deviation $\delta$ of backdoor attacks. 
Our theory categorizes defenses into two types: the \textbf{Hammer} (Type 2) and the \textbf{Anvil} (Type 1), based on whether they defend against small or large deviation attacks. From our theory, we derive that single-typed defenses and non-principled combined defenses are exploitable. We empirically validate this hypothesis under a new worst-case threat model: \textbf{the full-information adaptive adversary}, and we show that existing defenses are easily broken, answering our first research question (RQ 1). 

Our theory dictates the principled combination of defenses: combining Type 1 and Type 2 defenses to yield a robust principled combined defense, \textbf{\defenseshortname}. We introduce a new Type 2 defense: \textbf{Clipped-Super-Fine-Tuning (CSFT)}, and show that when combined with existing Type 1 defenses, it can even defend against our worst-case full-information adversary across a variety of data settings. We conjecture that combining specific Type 1 and Type 2 defenses yields a robust defense against all backdoor attacks, and we provide empirical evidence to support this claim (tentatively answering RQ 2). Our best combined defenses, $\text{HA}_{\text{Flame}}^{\text{CSFT}}$ and $\text{HA}_{\text{Multi-Metrics}}^{\text{CSFT}}$, are robust against both our worst-case full-information adversary and existing state-of-the-art attacks across all the data settings we evaluated.  

Extending our theory to other fields of machine learning security, such as adversarial robustness or watermarking, could introduce new research avenues. We leave some aspects of adaptive attack implementation and evaluation for future work. Additionally, future research into better Type 1 and Type 2 defenses would directly benefit our work, yielding even more robust combined defenses. 
Finally, reproducing our results across other data modalities, such as text or audio, could help further the understanding of federated learning backdoor attacks and their defenses. 

\begin{acks}
We gratefully acknowledge the support of NSERC for grants RGPIN-2023-03244, IRC-537591, the Government of Ontario and the Royal Bank of Canada for funding this research.
This paper was edited for grammar using Grammarly.
\end{acks}
\bibliographystyle{ACM-Reference-Format}
\bibliography{main}

\appendix

\section{Open Science}
To reproduce our results, the following are necessary:
\begin{itemize}
    \item The publicly available CIFAR-10 \cite{cifar10} and MNIST \cite{mnist} datasets.
    \item Our codebase, available at the following anonymous link: \url{https://anonymous.4open.science/r/Hammer-and-Anvil-0C21/}. It contains detailed instructions and code to reproduce our results, tables, and figures.
\end{itemize}

\section{Ethics considerations}
Our paper considers backdoor attacks and defenses against federated systems. While we propose a new, stronger attack, we also propose a defense that prevents it, thereby mitigating its impact. Our insights can also help further research into defenses against backdoor attacks on federated systems. Our datasets, models, defenses, and attacks are public and do not contain sensitive or private information. Our work did not use human subjects in any way.

\end{document}

%% file: tikz_figures/delta_gap.tex
\begin{figure}[htbp]
    \centering
    \begin{adjustbox}{width=0.99\columnwidth}
    \begin{tikzpicture}[node distance=2cm]
        
        % ==========================================
        % SCENARIO A: Top Row (Type 1 Defense Only)
        % ==========================================
        \begin{scope}[yshift=10.5cm]
            \node[anchor=south west, font=\large\bfseries] at (0, 1.8) {Scenario A: Type 1 Defense Only};
            \draw[thick, ->] (0,0) -- (10,0);
            
            % Axis Label
            \node[right] at (10,0) {$\delta$};
            \node[below left] at (10,0) {(Attack Magnitude)};
            
            % The Anvil
            \fill[blue!20, opacity=0.5] (6,0.2) rectangle (9.5,1.5);
            \node[blue!70!black, font=\normalsize, align=center] at (7.75,0.85) {Defense 1:\\Robust Aggregation\\or Outlier Detection};
            \draw[dashed] (6,0) -- (6,1.5);
            \node[below] at (6,0) {$\delta_1$}; 
            
            % Extended Gap (Left side) 
            \draw[->, thick] (0.1,0.5) -- (5.9,0.5);
            \node[font=\normalsize] at (3,0.8) {Exploitable Gap};
        \end{scope}

        % ==========================================
        % SCENARIO B: Middle-Top Row (Type 2 Only)
        % ==========================================
        \begin{scope}[yshift=7cm]
            \node[anchor=south west, font=\large\bfseries] at (0, 1.8) {Scenario B: Type 2 Defense Only};
            \draw[thick, ->] (0,0) -- (10,0);
            
            % Axis Label
            \node[right] at (10,0) {$\delta$};
            \node[below left] at (10,0) {(Attack Magnitude)};
            
            % The Hammer
            \fill[red!20, opacity=0.5] (0,0.2) rectangle (3,1.5);
            \node[red!70!black, font=\normalsize, align=center] at (1.5,0.85) {Defense 2:\\Removal};
            \draw[dashed] (3,0) -- (3,1.5);
            \node[below] at (3,0) {$\delta_2$}; 
            
            % Extended Gap (Right side) 
            \draw[<-, thick] (3.1,0.5) -- (9.5,0.5);
            \node[font=\normalsize] at (6.3,0.8) {Exploitable Gap};
        \end{scope}

        % ==========================================
        % SCENARIO C: Middle-Bottom Row (Combined)
        % ==========================================
        \begin{scope}[yshift=3.5cm]
            \node[anchor=south west, font=\large\bfseries] at (0, 1.8) {Scenario C: Principled Combined Defense};
            \draw[thick, ->] (0,0) -- (10,0);

            % Axis Label
            \node[right] at (10,0) {$\delta$};
            \node[below left] at (10,0) {(Attack Magnitude)};

            % The Hammer
            \fill[red!20, opacity=0.5] (0,0.2) rectangle (3,1.5);
            \node[red!70!black, font=\normalsize, align=center] at (1.5,0.85) {Type 2 Defense};
            \draw[dashed] (3,0) -- (3,1.5);
            \node[below] at (3,0) {$\delta_2$}; 

            % The Anvil
            \fill[blue!20, opacity=0.5] (6,0.2) rectangle (9.5,1.5);
            \node[blue!70!black, font=\normalsize, align=center] at (7.75,0.85) {Type 1 Defense};
            \draw[dashed] (6,0) -- (6,1.5);
            \node[below] at (6,0) {$\delta_1$}; 

            % Minimized Gap
            \draw[<->, thick] (3.1,0.5) -- (5.9,0.5);
            \node[font=\normalsize] at (4.5,0.8) {Exploitable Gap};
        \end{scope}

        % ==========================================
        % SCENARIO D: Bottom Row (Conjecture 1 - Overlap)
        % ==========================================
        \begin{scope}[yshift=0cm]
            \node[anchor=south west, font=\large\bfseries] at (0, 1.8) {Scenario D: Principled Combined Defense \& Conjecture 1 is True};
            \draw[thick, ->] (0,0) -- (10,0);
            
            % Axis Label
            \node[right] at (10,0) {$\delta$};
            \node[below left] at (10,0) {(Attack Magnitude)};

            % The Hammer (Extended further right to overlap)
            \fill[red!20, opacity=0.5] (0,0.2) rectangle (5.5,1.5);
            \node[red!70!black, font=\normalsize, align=center] at (2.5,0.85) {Strong Type 2 Defense};
            \draw[dashed] (5.5,0) -- (5.5,1.5);
            \node[below] at (5.5,0) {$\delta_2$}; 

            % The Anvil (Extended further left to overlap)
            \fill[blue!20, opacity=0.5] (4.5,0.2) rectangle (9.5,1.5);
            \node[blue!70!black, font=\normalsize, align=center] at (7.25,0.85) {Strong Type 1 Defense};
            \draw[dashed] (4.5,0) -- (4.5,1.5);
            \node[below] at (4.5,0) {$\delta_1$}; 

            % Notice: No gap arrow or text is drawn here, visually confirming Conjecture 1!
        \end{scope}

    \end{tikzpicture}
    \end{adjustbox}
    
    % Updated Caption to match the new 4-part layout
    \caption{Progression of the defensive spectrum against varying attack magnitudes ($\delta$). Scenario A and Scenario B highlight the exploitable gaps when relying on a single defense mechanism. Scenario C illustrates minimizing the security gap by combining mechanisms. Finally, Scenario D demonstrates Conjecture 1: an ideal state where the defenses overlap ($\delta_1 \leq \delta_2$), completely eliminating the exploitable gap.}
    \label{fig:security_gap}
\end{figure}

%% file: tikz_figures/attack_intuition.tex
\begin{figure*}[t]
    \centering

    % --- Subfigure A: Robust Aggregation ---
    \begin{subfigure}[b]{0.32\textwidth}
        \centering
        \begin{tikzpicture}[
            >=Stealth,
            benign/.style={circle, fill=blue!60, inner sep=1.5pt},
            malicious/.style={diamond, fill=red!70, inner sep=2pt},
            malicious_init/.style={diamond, fill=red!40, inner sep=2pt},
            center/.style={star, star points=5, fill=black, inner sep=1.5pt},
            font=\sffamily\small
        ]
            % Central model
            \node[center, label=below:{$w_{t}$}] (orig) at (0,0) {};
            
            % Clipping bound (eta)
            \draw[dashed, thick, gray] (0,0) circle (1.8cm);
            \node[gray, anchor=south west] at (1.27, 1.27) {Bound};
            
            % Benign points (scattered inside)
            \node[benign] at (0.3, 0.5) {};
            \node[benign] at (-0.6, 0.2) {};
            \node[benign] at (-0.2, -0.8) {};
            \node[benign] at (0.8, -0.3) {};
            \node[benign] at (0.5, 0.9) {};
            
            % Initial Malicious point (unscaled, inside the bound)
            \node[malicious_init, label=left:{$w_{m}$}] (m_init) at (-0.75, 0.5) {};
            
            % Malicious point scaled exactly to the bound
            \node[malicious, label=above left:{$w'_{m}$}] (m1) at (-1.5, 1.0) {};
            
            % Vector showing the origin to the initial point, then the scaling action
            \draw[dashed, thick, red!50] (orig) -- (m_init);
            \draw[->, thick, red] (m_init) -- (m1) node[midway, sloped, above, font=\scriptsize] {Scale};
        \end{tikzpicture}
        \caption{Robust Aggregation: The attacker scales the initial malicious weights ($w_m$) to rest exactly on the defense's acceptance bound, maximizing impact.}
        \label{subfig:robust_aggregation}
    \end{subfigure}
    \hfill
    % --- Subfigure B: Outlier Detection ---
    \begin{subfigure}[b]{0.32\textwidth}
        \centering
        \begin{tikzpicture}[
            >=Stealth,
            benign/.style={circle, fill=blue!60, inner sep=1.5pt},
            malicious/.style={diamond, fill=red!70, inner sep=2pt},
            center/.style={star, star points=5, fill=black, inner sep=1.5pt},
            font=\sffamily\small
        ]
            % Benign cluster (Faded Blue Background)
            \draw[fill=blue!5, draw=blue!30, dashed, thick] (0,0.5) circle (1.2cm);
            \node[blue!80] at (-0.8, 1.3) {Accepted};
            
            \node[benign] (bmean) at (0,0.5) {};
            \node[below] at (bmean) {$w_{b}$};
            \node[benign] at (0.3, 0.7) {};
            \node[benign] at (-0.2, 0.8) {};
            \node[benign] at (-0.4, 0.2) {};
            \node[benign] at (0.4, 0.1) {};

            % Ideal malicious update (far away)
            \node[malicious, fill=red!30, label=below:{\color{red!50}Ideal $w_{m}$}] (m_ideal) at (2.5, -1.0) {};
            
            % Line of bisection (Interpolation)
            \draw[thick, red!40, dashed] (bmean) -- (m_ideal);
            
            % Actual submitted malicious update (bisected to just cross the boundary)
            \node[malicious, label=above right:{$w'_{m}$}] (m_actual) at (1.0, -0.1) {};
            \draw[thick, red] (bmean) -- (m_actual);
        \end{tikzpicture}
        \caption{Outlier Detection: The attacker interpolates between the benign mean ($w_b$) and the ideal malicious weight, submitting the update for weights ($w'_m$) that just cross into the accepted cluster.}
        \label{subfig:outlier_detection}
    \end{subfigure}
    \hfill
    % --- Subfigure C: Removal Defenses ---
    \begin{subfigure}[b]{0.32\textwidth}
        \centering
        \begin{tikzpicture}[
            >=Stealth,
            benign/.style={circle, fill=blue!60, inner sep=1.5pt},
            malicious/.style={diamond, fill=red!70, inner sep=2pt},
            malicious_init/.style={diamond, fill=red!40, inner sep=2pt},
            center/.style={star, star points=5, fill=black, inner sep=1.5pt},
            font=\sffamily\small
        ]
            % Safe utility threshold (effective removal zone) - DRAWN FIRST
            \draw[fill=green!5, draw=green!60!black, dashed, thick] (0,0) circle (1.4cm);
            
            % Original benign model point - DRAWN SECOND (on top)
            \node[benign, label=below:{$w_{b}$}] (wb) at (0,0) {};
            
            % Label for the zone
            \node[green!60!black, anchor=south, fill=white, inner sep=0.5pt] at (0, 1.5) {Removal Zone};
            
            % Weak backdoor inside the zone
            \node[malicious_init, label=above left:{$w_{m}$}] (wm_in) at (0.6, 0.6) {};
            \node[green!60!black, font=\scriptsize, anchor=north] at (0.6, 0.5) {Removed};
            
            % Deeply embedded backdoor (scaled outside the zone)
            \node[malicious, label=right:{$w'_{m}$}] (wm_out) at (2.4, 1.6) {};
            \node[red, font=\scriptsize, anchor=north] at (2.5, 1.5) {Survives};
            
            % Vector showing the magnitude scaling
            \draw[->, thick, red] (wm_in) -- (wm_out) node[midway, sloped, above, font=\scriptsize] {Scale};
            
        \end{tikzpicture}
        \caption{Removal Defenses: A weakly embedded backdoor ($w_m$) falls within the effective removal zone and is removed. By scaling the update ($w'_m$) past this zone, the backdoor survives removal attempts.}
        \label{subfig:removal_defenses}
    \end{subfigure}

    \caption{A simplified representation of single-typed defenses and their associated adaptive attack principles. Blue circles represent benign weights, while red shapes denote adaptive malicious weights.}
    \label{fig:adaptive_attack_principles}
\end{figure*}

%% file: tables/lowest_m_table_badnet_train.tex
\begin{table}[ht]
  \centering
  \caption{Lowest number of malicious clients ($m$) out of 20 needed for our adaptive attacks to break existing defenses with the associated ASR and benign accuracies on CIFAR-10. We also include the $m = 8$ case. Values are averaged over three runs.}
  \label{tab:motivating_results}
  \begin{adjustbox}{width=0.99\columnwidth}
\begin{tabular}{l|ccc|cc}
\toprule
Defense & $m$ & ASR & Acc. & ASR ($m=8$) & Acc. ($m=8$) \\
\midrule
Multi Metrics & \textbf{4} & 98.86 & 76.16 & 99.30 & 78.61 \\
Flame & 2 & 55.61 & 82.16 & 98.25 & 80.06 \\
Rob MoM & 2 & 99.81 & 83.80 & 99.98 & 84.56 \\
MoM & 2 & 99.81 & 83.49 & 99.99 & 84.28 \\
Krum & 1 & 97.34 & 78.85 & 99.98 & 79.90 \\
Norm & 1 & 90.49 & 87.93 & 99.97 & 86.30 \\
CSFT & 1 & 87.47 & 79.01 & 98.70 & 82.15 \\
None & 1 & 99.62 & 83.02 & 99.99 & 84.64 \\
\bottomrule
\end{tabular}

  \end{adjustbox}
\end{table}

%% file: tables/lowest_m_simplified_table.tex
\begin{table}[t]
  \centering
  \caption{Lowest number of malicious clients ($m$) out of 20 needed for our adaptive attacks to break our HA defenses \textbf{after fine-tuning}, with the associated ASR and benign accuracies on CIFAR-10. >8 means we could not break the defense. Original $m$ values required to break the original defenses without HA are shown in parentheses. Values are averaged over three runs.}
  \label{tab:lowest_m_simplified}
\renewcommand{\arraystretch}{1.3}

  % \begin{adjustbox}{width=0.99\columnwidth}
\begin{tabular}{lccc}
\toprule
Defense & $m$ & ASR & Accuracy \\
\midrule
$\text{HA}_{\text{Flame}}^{\text{CSFT}}$ & \textbf{>8} (2) & \textbf{13.66} & 74.86 \\
$\text{HA}_{\text{Krum}}^{\text{CSFT}}$ & \textbf{>8} (1) & 26.87 & 76.03 \\
$\text{HA}_{\text{Multi-Metrics}}^{\text{CSFT}}$ & \textbf{>8} (4) & 35.51 & 79.56 \\
$\text{HA}_{\text{Norm}}^{\text{CSFT}}$ & 4 (1) & 67.86 & \textbf{83.50} \\
$\text{HA}_{\text{MoM}}^{\text{CSFT}}$ & 2 (2) & 84.73 & 79.11 \\
$\text{HA}_{\text{Rob-MoM}}^{\text{CSFT}}$ & 2 (2) & 94.26 & 81.18 \\
\bottomrule
\end{tabular}

  % \end{adjustbox}
\end{table}

%% file: tables/existing_attacks_m8_table.tex
\begin{table}[t]
  \centering
  \caption{Evaluation of \defenseshortname~against attacks from previous work. We report the ASR and benign accuracies (Acc.) for $m=8$.}
  \label{tab:existing_attacks_m8}
  \begin{adjustbox}{width=0.99\columnwidth}
  \renewcommand{\arraystretch}{1.3}
\begin{tabular}{l|cc|cc|cc}
\toprule
Defense & \multicolumn{2}{c|}{$\text{HA}_{\text{Krum}}^{\text{CSFT}}$} & \multicolumn{2}{c|}{$\text{HA}_{\text{Flame}}^{\text{CSFT}}$} & \multicolumn{2}{c}{$\text{HA}_{\text{Multi-Metrics}}^{\text{CSFT}}$} \\
Attack & ASR & Acc. & ASR & Acc. & ASR & Acc. \\
\midrule
DBA & 3.61 & 72.50 & 3.77 & 77.37 & \textbf{3.84} & \textbf{78.74} \\
BadNet & 4.04 & 74.37 & 3.33 & \textbf{78.24} & 3.10 & 78.56 \\
Neurotoxin & 3.96 & 74.52 & 3.18 & 78.21 & 3.20 & 77.79 \\
Replacement & \textbf{21.69} & \textbf{74.80} & \textbf{3.93} & 77.62 & 3.30 & 78.05 \\
\bottomrule
\end{tabular}

  \end{adjustbox}
\end{table}

%% file: tables/lowest_m_simplified_non_iid.tex
\begin{table}[t]
  \centering
  \caption{Lowest number of malicious clients ($m$) out of 20 needed to break our HA defenses on non-IID CIFAR-10 (Dirichlet $\alpha=0.5$). Original $m$ values are in parentheses. Values are averaged over three runs.}
  \label{tab:lowest_m_simplified_non_iid}
  \renewcommand{\arraystretch}{1.3}
  % \begin{adjustbox}{width=0.99\columnwidth}
\begin{tabular}{lccc}
\toprule
Defense & $m$ (Original) & ASR & Accuracy \\
\midrule
$\text{HA}_{\text{Flame}}^{\text{CSFT}}$ & \textbf{>8} (4) & \textbf{12.63} & 70.14 \\
$\text{HA}_{\text{Krum}}^{\text{CSFT}}$ & \textbf{>8} (1) & 19.94 & 73.32 \\
$\text{HA}_{\text{Multi-Metrics}}^{\text{CSFT}}$ & \textbf{>8} (2) & 37.89 & \textbf{74.31} \\
\bottomrule
\end{tabular}

  % \end{adjustbox}
\end{table}

%% file: tables/lowest_m_simplified_mnist.tex
\begin{table}[t]
  \centering
  \caption{Lowest number of malicious clients ($m$) out of 20 needed to break our HA defenses on the MNIST dataset. Original $m$ values are in parentheses. Values are averaged over three runs.}
  \label{tab:lowest_m_simplified_mnist}
  \renewcommand{\arraystretch}{1.3}
  % \begin{adjustbox}{width=0.99\columnwidth}
\begin{tabular}{lccc}
\toprule
Defense & $m$ (Original) & ASR & Accuracy \\
\midrule
$\text{HA}_{\text{Flame}}^{\text{CSFT}}$ & \textbf{>8} (8) & 21.79 & 97.73 \\
$\text{HA}_{\text{Multi-Metrics}}^{\text{CSFT}}$ & \textbf{>8} (4) & 37.12 & 97.56 \\
$\text{HA}_{\text{Krum}}^{\text{CSFT}}$ & \textbf{>8} (1) & 11.68 & 97.80 \\
\bottomrule
\end{tabular}

  % \end{adjustbox}
\end{table}

%% file: tables/optimal_early_stopping_table.tex
\begin{table}[t]
  \centering
  \caption{Comparison of early-stopping epoch timings during fine-tuning. 'Safe' epoch reports the epoch that maximizes benign accuracy while successfully keeping ASR strictly below 50\%. 'Peak' epoch reports the epoch with the absolute highest benign accuracy regardless of the backdoor ASR. Both strategies mandate a minimum of 10 fine-tuning epochs.}
  \label{tab:optimal_early_stopping}
  \begin{adjustbox}{width=0.99\columnwidth}
    \renewcommand{\arraystretch}{1.3}

\begin{tabular}{l|ll|ll|ll|ll}
\toprule
\hfill m \qquad & \multicolumn{2}{c|}{1} & \multicolumn{2}{c|}{2} & \multicolumn{2}{c|}{4} & \multicolumn{2}{c}{8} \\
Defense & Safe & Peak & Safe & Peak & Safe & Peak & Safe & Peak \\
\midrule
$\text{HA}_{\text{Flame}}^{\text{CSFT}}$ & 98 & 98 & 94 & 94 & 93 & 93 & 96 & 96 \\
$\text{HA}_{\text{Krum}}^{\text{CSFT}}$ & 97 & 97 & 99 & 99 & 100 & 100 & 100 & 100 \\
$\text{HA}_{\text{Multi-Metrics}}^{\text{CSFT}}$ & 100 & 100 & 94 & 94 & 91 & 91 & 97 & 97 \\
\bottomrule
\end{tabular}

  \end{adjustbox}
\end{table}